# Adapting Vision-Language Models for Human-Readable XAI in Industrial Object Detection

Sarvenaz Sardari[a,b,*], Freddy Fernandes[a], Samarth Yelvande[a], Jose Moises Araya-Martinez [a], Alina Roitberg[b]

[a]*Mercedes-Benz AG, Future Automotive Manufacturing, Benz-Str. Bau40, 71063 Sindelfingen, Germany*
[B]*University of Stuttgart, AI, Intelligent Sensing and Perception, Universitätsstr. 38, 70569 Stuttgart, Germany*

* Corresponding author. *E-mail address:* sarvenaz.sardari@mercedes-benz.com

**Abstract**

Explainable Artificial Intelligence (XAI) solutions are essential for building trust in AI technologies and their integration in real manufacturing lines. However, most existing methods are tailored to technical experts, limiting their accessibility to diverse user groups such as blue-collar workers in manufacturing lines who use AI for quality control. In this work, we introduce an XAI interface for object detection in industrial manufacturing based on a fine-tuned vision-language model, designed to generate intuitive explanations for non-expert users. We benchmark existing vision-language models and demonstrate that out-of-the-box models often fall short in delivering clear, context-relevant explanations for non-expert users. To address this, we fine-tune a vision-language model and integrate it into our interface, enabling contextualized, accessible explanations for non-expert users. We demonstrate improvements in explanation clarity, instruction adherence, image groundedness, and contextual awareness over GPT 4o-mini on proprietary and public robotics dataset. This approach advances the accessibility and usability of AI explanations, making them more intuitive and applicable in manufacturing domain.

## 1. Introduction

The automotive industry is increasingly adopting Artificial Intelligence (AI) in production lines. However, the black-box nature of many AI models makes it difficult for them to be trusted in high-stakes applications, such as quality inspection [1]. In these contexts, errors can result in environmental, material, and financial costs [2]. The AI Act of the European Union mandates that the design of high-risk AI systems enables deployers to “interpret a system’s output” [3]. Because of the explainability-performance trade-off, transparent models often fall short of the performance required for industrial applications [4]. As a result, post hoc explainability methods are needed to interpret existing high-performance models. These techniques offer insight into why a particular decision was reached without altering the original architecture [5]. However, these techniques are often not intuitive or accessible to non-experts, as they are primarily developed by and for technical specialists [6,7]. Large language models (LLMs) can serve as a bridge between complex eXplainable AI (XAI) outputs and end users, making explanations more understandable and better suited for domain experts across different fields [8,9]. Nevertheless, the unique requirements of the automotive industry such as data confidentiality and strict regulatory constraints limit the adoption of existing large vision-language models (VLMs), particularly those that are not open-source [10]. Customizing large VLMs for specific downstream tasks typically requires substantial computational resources [11]. Parameter Efficient Fine-Tuning (PEFT) offers a practical alternative by adjusting the parameters of a pre-trained model to adapt it to a specific task or domain [11]. Following fine-tuning, it is essential to evaluate the model's ability to generate suitable explanations using well-defined explainability metrics [12]. In this study, we benchmark existing pre-trained language models using explanation-specific metrics tailored to the industrial domain. Our results indicate that while general-purpose models show promise, they are not yet suitable for explaining industrial AI models out of the box. To address this, we introduce a customized, domain-specific, VLM to translate complex object detection explanations into intuitive insights for end users following the pipeline in Fig. 1. We demonstrate the effectiveness of our approach by testing the model on the publicly available robotics dataset [13].

In summary, our work introduces the following key contributions:

- Benchmarking different VLMs for generating human understandable XAI explanation for object detection.
- Releasing an open source, custom vision language model tailored for user centric explanations in industrial object detection applications [14].
- Proposing a training pipeline for developing domain-specific VLMs for XAI.

In the following sections, we review work related to our approach. We then show details about our methodology and implementation, followed by a summary of the most relevant results on proprietary and public datasets. In the last section, we draw our conclusions and propose potential future work.

## 2. Related work

LLMs have recently been leveraged to make AI explanations more accessible to non-experts. Applications span diverse domains: in healthcare, LLMs help build trust in AI-driven diagnostics [15]; in finance, they add transparency to credit scoring and fraud detection [16]; and in manufacturing, they clarify decisions in predictive maintenance by linking model predictions to input features like vibration or temperature [17,18]. However, these studies typically focus on textual data, leaving a gap in vision-based explainability. XEdgeAI [19] and LangXAI [20] has been introduce to incomprate feauture based XAI and LLMS to deliver human-centered interpretability through visual and textual explanations to end-users. XEdgeAI relies on ground truth image as input which makes it not suitable for real time inferences and LangXAI is limited due to domain mismatch.

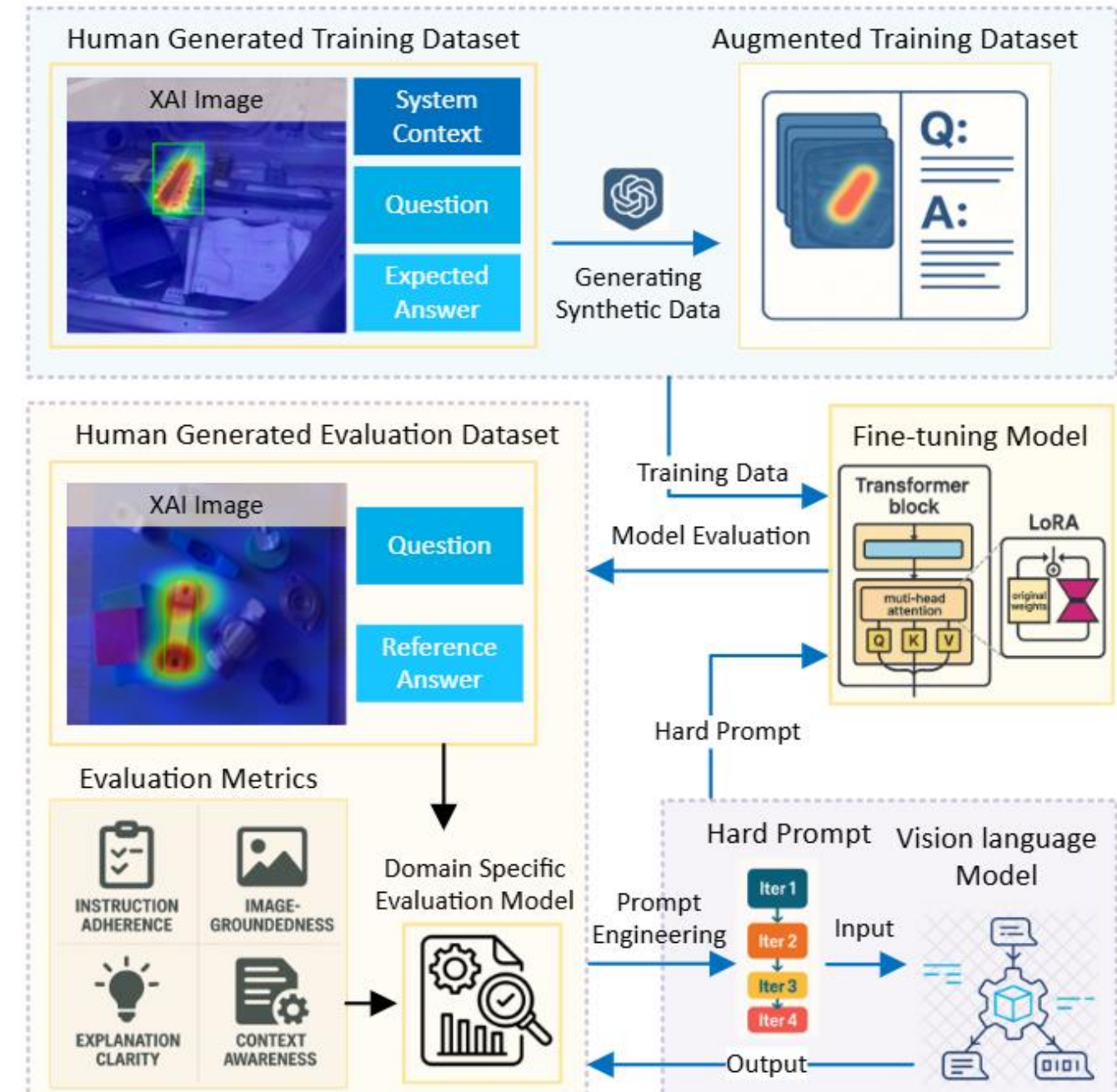

Fig. 1. Vision-language-model fine-tuning pipeline consisting of 4 main parts of data generation, prompt engineering, fine-tuning and evaluation.

To apply VLMs to domain-specific tasks like industrial object detection, fine-tuning is essential, often requiring synthetic data when real-world datasets are scarce. Prior work has shown that combining real and synthetic data improves performance [21], Maheshwari et al. propose a four-part framework for generating such data: task descriptions, few-shot examples, data generation guidelines, and output formatting [22]. Using multiple LLMs to generate synthetic data reduces bias and increases robustness [23]. To the best of our knowledge, there are currently no benchmarking datasets focused on image-based explanations in industrial or contexts while existing benchmarks are largely text-based [24].

Evaluation of fine-tuned models for human-centered XAI remains challenging. There is no consensus on which explanation properties are most meaningful to users [25]. Zhou et al. propose metrics bridging LLM and XAI evaluation [12], which we adapt with expert input for industrial use cases. While GPT-4 [26] and similar models are often used as evaluators [27], their closed-source nature, excessive costs, and lack of version control pose limitations [28]. To address this, we adopt Prometheus, an open-source evaluation framework by Kim et al., aligned with our domain-specific rubrics [29].

## 3. Methodology and Experimental Set Up

This section outlines the steps taken to develop and evaluate our VLM for industrial object detection human-centered XAI. We describe the dataset creation, model fine-tuning, and evaluation setup in the following subsections.

### 3.1. Model Selection

The model selection process was guided by a focus on open-sourced, fine-tuneable vision-language architectures. Emphasis was placed on models that are both compact (under 10B parameters), facilitating deployment on edge devices or standard cloud instances, and representing recent advancements in the field. This results in developing smaller, yet powerful, language and vision models that offer a balance of performance and accessibility [30]. From this candidate pool, we prioritized models with leading benchmark performances and robust community support, ensuring both technical quality and practical usability. Specific exclusions were necessary to avoid potential methodological bias, LLaVA-based models [31], were omitted due to underpinning the Prometheus evaluation model. This systematic filtering led to the selection of Qwen2.5-7B [32], InternVL3-8B [33], and Idefics3-8B [34]. Each of these models meets our criteria on relevant multi modal tasks.

### 3.2. Dataset Creation

To evaluate and subsequently fine-tune our selected VLMs, we developed an evaluation dataset and a more extensive training dataset.

For assessing model performance on specific XAI use cases, we created evaluation datasets comprising of 50 unique instances. Each instance contains an XAI visualization image containing saliency map generated from D-RISE [35] and detected object bounding box, a refined question related to the visualization from general XAI question bank published by Vera et al. [36], and a human-generated reference answer. This human-authored ground truth is important for metrics that assess alignment with human reasoning and interpretation.

For the fine-tuning phase, we created a larger training dataset with a similar structure to the evaluation dataset, incorporating image-question pairs enhanced by specific system context prompts. For each training dataset instance, we generated synthetic images from manufacturing parts used by Araya-Martinez et al. [37]. Dataset creation involved a two-phase approach:

1- Initial seed: Domain experts manually curated an initial seed set of approximately 150 high-quality examples.

2- Data augmentation: This seed set was then expanded through synthetic data augmentation, employing OpenAI's BatchAPI with the GPT-4o Mini model, selected for its balance between performance and cost-effectiveness.

The final augmented dataset comprised approximately 1300 instances. Throughout this augmentation, careful attention was given to maintaining data quality and relevance, mitigating potential inaccuracies or stylistic deviations typical of synthetic data generation.

### 3.3. Fine-tuning

We employed hard prompt engineering, a model-agnostic process that maintains fixed model weights while iteratively refining the prompt text. Conducted over seven iterations as shown in Table 1, this strategy progressively enhanced the clarity, structure, and overall quality of the VLM outputs. This approach aligns with established prompt engineering strategies for optimizing model performance through principled instruction design [38]. The prompt evolution was as follows:

- Iteration 1, initial prompt: Established a baseline with fundamental context, task, and a general objective for explaining saliency map visualizations.
- Iteration 2, chain-of-thought: Incorporated step-by-step instructions to guide the model's reasoning process, a technique aligned with Chain-of-Thought prompting.
- Iteration 3, explicit output formatting: Added clear directives for a single-sentence response (10-20 simple words), targeting non-technical users and avoiding jargon, to improve conciseness and suitability.
- Iteration 4, bullet instructions: Converted instructions to bullet points to enhance prompt structure and model focus on key aspects.
- Iteration 5, few-shot example: Introduced an example response, demonstrating the desired style and content as a form of few-shot prompting or in-context learning.
- Iteration 6, Contextual Refinement: Restructured and enhanced the "Context" section with clearer definitions for "Bounding Box" and "Saliency Map," while retaining the example.
- Iteration 7, Example Elimination: Removed the example response to prevent potential overfitting to its phrasing and to grant the model more generative freedom. This final iteration yielded the best performance scores.

This iterative refinement highlighted the importance of balancing detailed guidance with sufficient flexibility, leading to a final prompt that effectively elicited the desired high-quality, structured outputs.

Table 1. Hard-prompts benchmark results (Prometheus score, 1-5).

| Prompts | Idefics3 | Qwen2.5 | Intern3 |
|---|---|---|---|
| Iteration 1 | 2.09 | 3.85 | 4.57 |
| Iteration 2 | 2.45 | 4.04 | 4.43 |
| Iteration 3 | 2.68 | 2.80 | 3.30 |
| Iteration 4 | 2.76 | 2.95 | **3.57** |
| Iteration 5 | 2.52 | 2.91 | 3.26 |
| Iteration 6 | 2.53 | 2.92 | 2.93 |
| Iteration 7 | **2.81** | **3.19** | 3.33 |

To enhance model performance without the need for complete model retraining, we employ Adapter Tuning using Low-Rank Adaptation (LoRA) [39]. LoRA achieves this by injecting compact, trainable low-rank matrices into strategically selected layers of the transformer architecture, specifically targeting only a small portion of the overall parameters. In all three of our ~7–8 B-parameter VLMs we adapt the same seven projection layers. The targeted modules encompass several critical projection layers within the transformer architecture, including the query, key, value, and output projections within the self-attention mechanisms, as well as projections in the feed-forward networks such as the gate projection, up-projection, and down-projection.
These layers are the most leverage-rich targets for parameter-efficient tuning. They collectively decide what information is attended to, how it is combined across modalities, and which transformed features flow forward. Early experiments that restricted LoRA to language-only projections (i.e., omitting the vision-critical projections above) produced minimal gains on image grounding, context awareness, or instruction adherence.

### 3.4. Evaluation Method

For continuous evaluation, we adopted an automated approach using Prometheus 13B [30]. This open-source LLM, with evaluation capabilities comparable to GPT-4, allows for the use of reference answers facilitating a robust assessment of our vision-language models' correctness and reliability. Our evaluation framework is based on four primary scoring rubrics, with detailed scores presented in Fig 2 and Fig 3:

- Explanation Clarity: Assesses how clear and understandable the model's explanation is for the intended audience.
- Image-Groundedness: Measures the relevance and accuracy of the response in referencing visual content.
- Context-Awareness: Evaluates the use of relevant external information to enhance the explanation.
- Instruction Adherence: Gauges how closely the model's output follows the provided instructions.

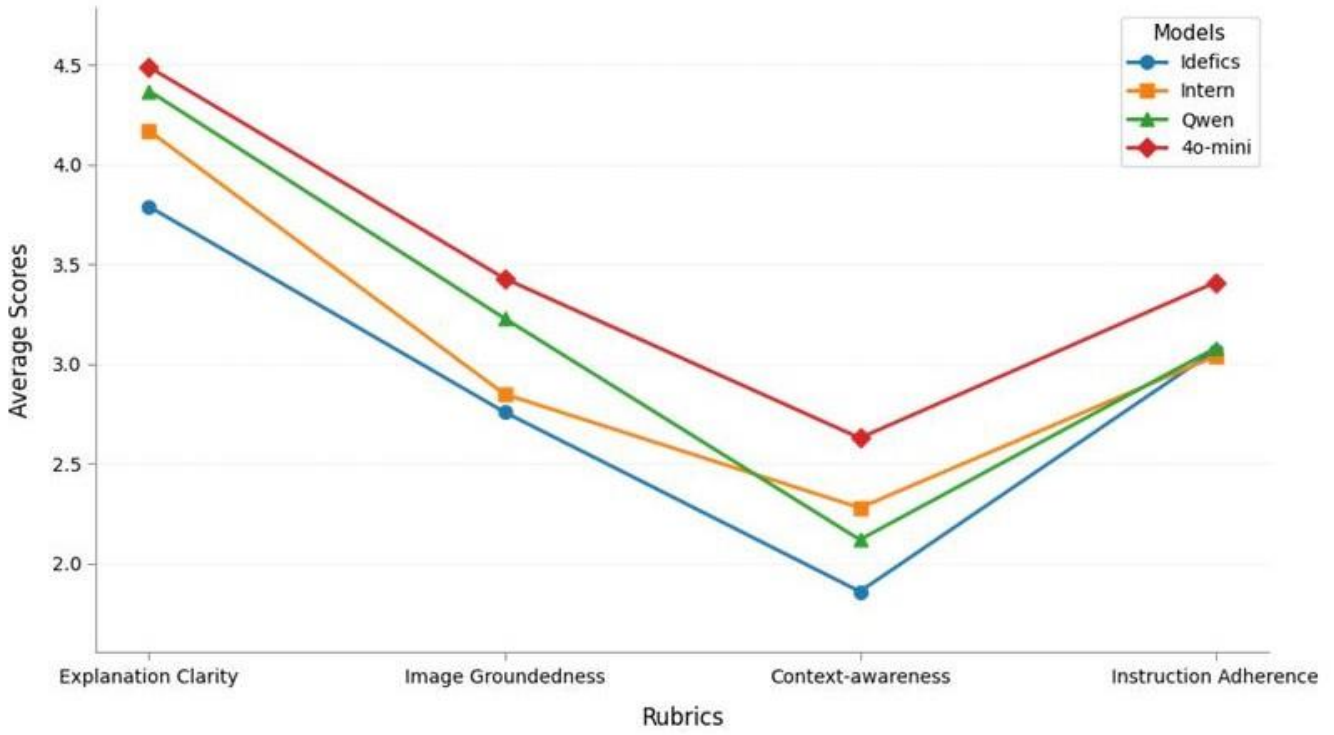

Fig. 2. Evaluation of vanilla VLMs. While models demonstrate strong general-language capabilities, their performance declines in tasks requiring domain-specific understanding and visual-context alignment.

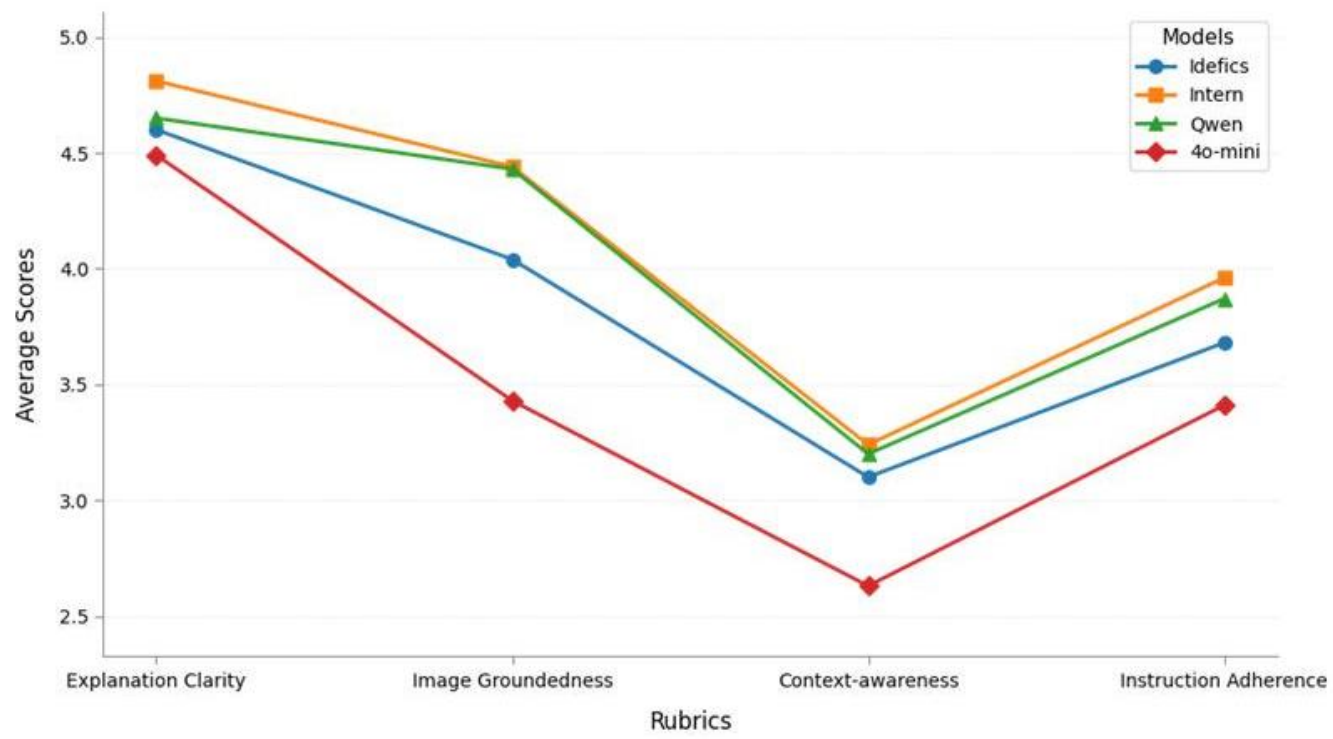

Fig. 3. Post-fine-tuning evaluation scores of the models. Improvements are observed for Idefics and Qwen. Intern shows minimal gains, highlighting model-dependent variability in fine-tuning efficacy.

Beyond these rubrics, response length was a critical metric. In industrial settings, overly verbose outputs may be disregarded by end users and can slow down decision-making processes. Consequently, from Hard prompt iteration 3 onwards, we incorporated stricter instructions specifying the desired concise response length to better align outputs with practical user needs.

## 4. Results and Discussion

This section presents results across three areas: hard prompt benchmarking, model performance before and after fine-tuning, and evaluation on a public robotics dataset.

### 4.1. Hard prompts Benchmarking

The results from our iterative hard prompt engineering indicate that Qwen and Intern achieve higher scores during the initial two iterations. However, without imposing a maximum token limit on outputs, the explanations produced by these models frequently become excessively long, typically ranging between 50 and 200 words, which reduces their suitability for production use. While setting a maximum token limit can address this issue by ensuring more concise and complete outputs, it also carries the risk of prematurely cutting off explanations. Introducing examples into the prompt led to a notable drop in performance. This is primarily because the models tend to inappropriately reuse features from the example rather than adhering to the intended output structure, sometimes producing repetitive or off-topic responses. While the final iteration does not yield the highest overall scores as general use cases, it highlights the latent potential of these models to improve substantially through fine-tuning.

### 4.2. Vanilla and Fine-tuned Models Performance

As shown in Fig. 2, the vanilla models are already linguistically fluent: their Explanation Clarity ranges from 3.8 to 4.4 out of 5, indicating that generated answers are understandable to non-expert users. This strength, however, reflects general language proficiency rather than domain understanding. Performance drops sharply on vision-specific criteria since dataset contains more domain specific objects. Image Groundedness lies between 2.8 and 3.2, signalling limited ability to tie statements to concrete regions in the image, while Context-awareness remains below 2.3, revealing difficulty in incorporating scene context and background knowledge. Instruction Adherence hovers around 3.0, suggesting only partial compliance with task-specific directives. These deficiencies motivate targeted fine-tuning.

After applying parameter-efficient LoRA adapters to every projection sub-module, all three models improve across all rubrics as shown in Fig. 3. The largest gains appear where baseline performance was weakest. Image Groundedness rises by +1.3 to +1.6 points, pushing every model above 4.0. Context-awareness climbs by just over one point, showing that the models now weave scene details and domain knowledge into coherent explanations. Instruction Adherence approaches the 4 point, with noticeably fewer prompt omissions and less over-generalisation. Explanation Clarity records the smallest absolute gain—its baseline was already high—but still advances to 4.6–4.8, demonstrating that richer visual reasoning enhances, rather than compromises, readability.

Fig. 4 shows these fine-tuned models against GPT-4o Mini (mean score ≈ 3.5). Whereas the open-source models initially lagged behind, the LoRA-adapted versions of Idefics, Qwen, and Intern now all exceed the proprietary baseline. These results show that injecting a small number of trainable parameters—less than 1% of the model—can unlock multimodal competence that matches or surpasses a commercial counterpart of comparable size in domain specific tasks, underscoring the practical value of parameter-efficient fine-tuning for industrial vision-language applications.

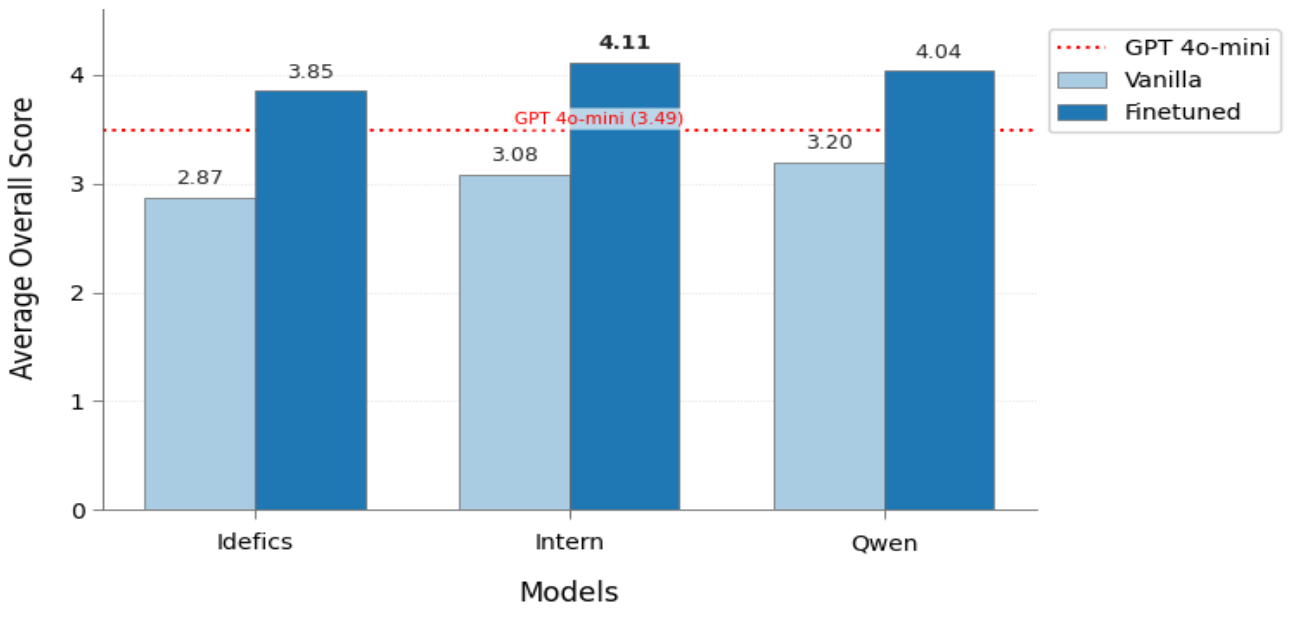

Fig. 4. Comparison of performance between GPT-4o Mini and fine-tuned models, indicating that fine-tuned models outperform GPT-4o Mini.

### *4.3. Performance on Public Robotic Dataset*

To assess the generalizability of our fine-tuned VLMs across diverse industrial and robotic object detection tasks, we evaluated its performance on the publicly available Sim2Real dataset [14], which features different object categories and scene layouts which are not present in the training manufacturing parts dataset. The results are summarized in Table 2 as well as sample explanations shown in Fig.5. Qwen continues to improve, achieving an overall score of 4.17. Idefics and Intern show slight drops yet remain strong with high scores throughout, respectively. These findings suggest that the LoRA-tuned models retain robustness across applications and are suitable for a broader range of real-world robotic applications.

Table 2. Final score of finetuned models on public dataset.

| Criteria | Idefics3 | Qwen2.5 | Intern3 |
|---|---|---|---|
| Explanation Clarity | 4.44 | 4.62 | 4.62 |
| Instruction Adherence | 3.64 | 3.98 | 3.76 |
| Image Groundedness | 3.94 | 4.46 | 4.14 |
| Context-awareness | 3.02 | 3.64 | 3.06 |
| Overall performance | 3.76 | **4.17** | 3.90 |

## 5. Conclusion and Future Work

In this study, we addressed the critical need for accessible and human-centered explanations in industrial AI applications by fine-tuning a VLM to generate post-hoc explanations for object detection tasks. Our evaluation demonstrates that, while vanilla VLMs exhibit strong linguistic capabilities, they fall short in domain-specific tasks. Through targeted fine-tuning on a carefully designed dataset, our models significantly improve across these rubrics, outperforming even closed-source models like GPT-4o Mini in industrial relevance. Furthermore, we show that the fine-tuned model generalizes effectively to other object detection scenarios, as evidenced by consistent performance on the Sim2Real dataset, indicating its broader applicability in robotics and manufacturing contexts.

Future work will focus on extending this approach for application in classification models across various industry domains. Investigate multi-modal fine-tuning that includes the vision backbone, improving visual grounding and external knowledge integration. Additionally, we aim to explore interactive explanation frameworks and adaptive prompting strategies to further personalize the explanatory output for diverse end-user profiles in industry.

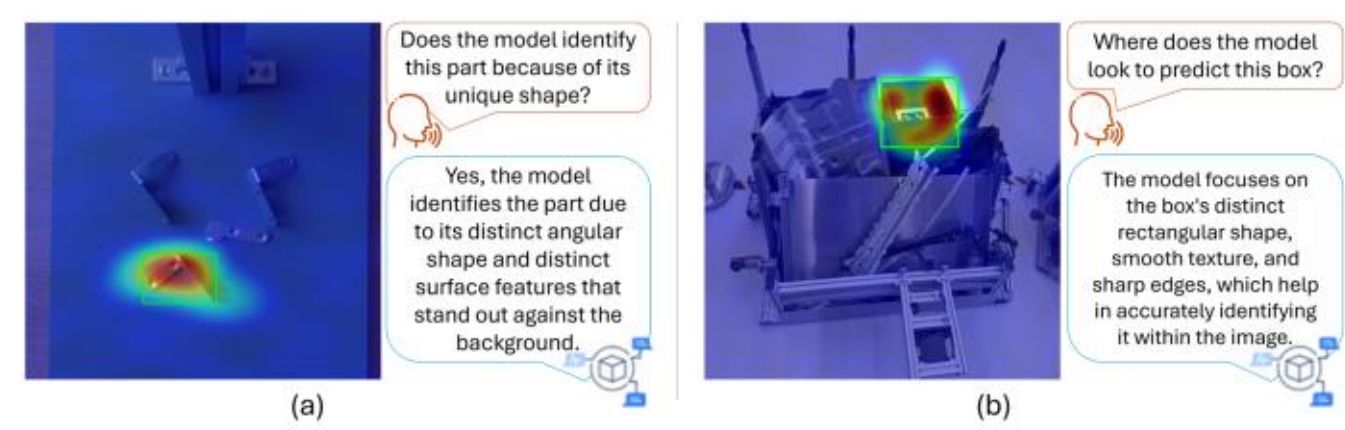

Fig. 5. Representative examples of explanations for (a) public robotic dataset and (b) manufacturing parts dataset.

## Acknowledgements

This work has been funded by the German Federal Ministry for Economic Affairs and Climate Action based on a resolution of the German Bundestag, financed by the European Union.